\documentclass[11pt,a4paper]{article}
\usepackage[utf8]{inputenc}
\usepackage[T1]{fontenc}
\usepackage{times}
\usepackage[margin=1in]{geometry}
\usepackage{amsmath,amssymb}
\usepackage{booktabs}
\usepackage{graphicx}
\usepackage{xurl}
\usepackage[hidelinks]{hyperref}
\usepackage[round,authoryear]{natbib}
\usepackage{xcolor}
\usepackage{array}
\usepackage{enumitem}
\usepackage{framed}
\usepackage{caption}
\setlist{nosep,leftmargin=1.5em}

\title{Screen Before You Interpret:\\A Portable Validity Screen for Benchmark-Based LLM Confidence Signals}
\author{Jon-Paul Cacioli\\
Independent Researcher, Melbourne, Australia\\
ORCID: 0009-0000-7054-2014}
\date{Version 2 (corrected and consolidated), November 2026}

\begin{document}
\maketitle

\begin{abstract}
\noindent LLM confidence signals drive abstention, routing and escalation, yet evaluations rarely check whether a signal varies with correctness before computing metrics on it. Clinical personality assessment handles the analogous problem by screening response validity before interpreting substantive scales. We specify a portable screen in that tradition. From the correctness-by-confidence table it computes the rate of high confidence on errors (L), the rate of low confidence on correct answers (Fp), and their contrast (RBS, equal to minus Youden's $J$). It adds no new statistics: it separates blanket confidence, blanket withdrawal and inversion, attaches interval guards, and calls a signal Valid only when it measurably tracks correctness. On 20 frontier models answering 452 two-alternative items with binary KEEP/WITHDRAW probes, one model is Invalid (DeepSeek-R1 withdraws 95\% of its correct answers), four are Indeterminate and 15 are Valid. Two Indeterminate models keep confidence on 96\% of their errors but make too few errors to certify it; certification needs 35 to 139 errors. Only 6 of the 20 verdicts reproduce reliably on half the items. On verbalized 0--100 confidence (33 models on 1,500 MMLU items, 20 on a second 500-item sample, and Yang et al.'s data for 11 models on 10 benchmarks), the verdict depends on the cut point: at 50 most models are Invalid because answers pile at the ceiling, while at 90--95 most pass. Battery item sensitivity does not predict MMLU discrimination ($\rho = -.07$). Protocol amendments were settled on these data, so the verdicts are descriptive pending a held-out test. This version corrects scoring, grading, classification and reporting errors in version 1 and supersedes arXiv:2604.17707 and 2604.17716.
\end{abstract}

\begin{framed}
\noindent\textbf{Correction and consolidation notice (v2).} Version 1 of this paper and its two companions (arXiv:2604.17707, \emph{Before You Interpret the Profile}; arXiv:2604.17716, \emph{Concurrent Criterion Validation}) contained errors that change headline results. This version replaces all three. (1) The battery's free-text track recorded 71 of DeepSeek-R1's 72 responses as declines, with the harness default of KEEP+BET, and scored them as errors; this produced an artefactual ``inverted monitoring'' profile for R1 (reported $r = -.80$). (2) The same track was graded by string matching that marks correct answers wrong and wrong answers correct; it is excluded from the primary analysis (Section~\ref{sec:data}). (3) The reported tiers were not produced by the protocol as implemented; this version reports the tiers that the released tool (\texttt{validity\_screen} 0.3.0) produces, with the amendments in Appendix~\ref{app:amend}. (4) The selective-prediction ``criterion validation'' used a criterion computed from the same items and probe as the screen; it is reported here descriptively. (5) The v1 reanalysis of Yang et al.\ (2024) reported results that cannot be reproduced from that study's released data; it is replaced by an analysis of the released item-level predictions. Appendix~\ref{app:corrections} lists the changed claims, and Section~\ref{sec:upstream} states what this means for two papers that share the data.
\end{framed}

\section{Introduction}

When a language model states how confident it is, downstream systems act on that statement. Abstention systems withhold low-confidence answers \citep{wen2025}, routers send uncertain queries elsewhere, and safety pipelines escalate to human review. The calibration literature asks how well stated confidence matches accuracy \citep{xiong2024,geng2024,yang2024}; the metacognition literature asks whether models can tell what they know from what they do not \citep{kadavath2022,steyvers2025}; the selective-prediction literature asks how much reliability abstention can buy \citep{geifman2017}. All three take a prior condition for granted: that the confidence signal varies with correctness at all.

That condition often fails in recognisable ways. A model can report high confidence on nearly every item, so that its confidence on errors cannot be told apart from its confidence on correct answers. It can do the opposite and withdraw nearly everything. Aggregate metrics register both as a low or uninformative score without saying which pattern produced it, and they are usually reported without first asking whether the signal was interpretable.

Clinical personality assessment met this problem decades ago. Instruments such as the MMPI-3 \citep{benporath2020mmpi3} and the PAI \citep{morey1991,morey2007} contain validity scales that are checked before any clinical scale is interpreted. A respondent who answers ``true'' to every item, or denies every problem, produces clinical scores that reflect a response set rather than item content; when the validity scales exceed their thresholds, the clinical scores are not interpreted \citep{vanscoyoc2017,burchett2022}. The data structure is the same as in LLM confidence evaluation: binary responses to structured items, where the question is whether the pattern carries item-level information or is dominated by a fixed way of responding.

This paper specifies a portable validity screen for benchmark-based LLM confidence data in that tradition and evaluates it on four data sets. For a binary signal its indices are standard quantities: the two conditional error rates of the correctness-by-confidence table, minus Youden's $J$ \citep{youden1950}, and the $\phi$ coefficient. Its contribution is not new statistics but (i) a decomposition that separates blanket confidence, blanket withdrawal and inversion, which call for different responses; (ii) an ordered convention with interval guards, an explicit ``cannot certify'' outcome, and a Valid verdict that requires evidence of item sensitivity; (iii) a reporting table; and (iv) an account of when verdicts are and are not stable, including a sample-size requirement and a dependence on the cut point used to binarise continuous confidence.

\paragraph{Research questions.}
\begin{itemize}
\item \textbf{RQ1.} How does the released implementation classify response policies whose properties are known?
\item \textbf{RQ2.} On binary probes, which response sets does the screen find in frontier models, and what does it add to a direct test of item sensitivity?
\item \textbf{RQ3.} How stable are verdicts, and how many errors does a verdict require?
\item \textbf{RQ4.} How does the screen behave on verbalized 0--100 confidence, and do verdicts carry over between probe formats?
\end{itemize}

\paragraph{What the mapping does and does not claim.} The transfer from clinical assessment is structural. We do not claim that models have personalities, experience doubt or manage impressions. Signal detection theory was developed for human observers, yet $d'$ measures sensitivity for any detector \citep{green1966}; the validity logic transfers in the same way. Clinical constructs, normative reference populations and motivational interpretations do not transfer. We keep the clinical index names because the companion literature uses them, and give the standard name for each in Table~\ref{tab:indices}.

\section{The Screen}\label{sec:protocol}

\subsection{Two stages and the inputs}

The screen (Stage A) runs before any substantive confidence metric (Stage B: AUROC, calibration error, meta-$d'$, risk--coverage curves, selective accuracy). It needs a set of items with binary correctness labels and one confidence judgement per item that is binary or can be binarised. Items the model did not answer (an explicit decline, or output the parser could not map to an answer) are excluded before scoring and counted, as are confidence values that the pipeline rather than the model supplied, such as a parser's default. A declined item carries no confidence judgement about an answer; scoring it as a confident error, as version 1 did, manufactures a response set the model did not produce.

The scored items form one table per model:
\begin{center}
\begin{tabular}{lcc}
\toprule
 & Correct & Incorrect \\
\midrule
High confidence (KEEP) & $a$ & $b$ \\
Low confidence (WITHDRAW) & $c$ & $d$ \\
\bottomrule
\end{tabular}
\end{center}

\subsection{Indices}

\begin{table}[t]
\centering
\caption{Indices, definitions and standard equivalents for a binary signal. $n_K$, $n_W$: numbers of high- and low-confidence responses; $N$: scored items.}
\label{tab:indices}
\small
\begin{tabular}{l p{2.9cm} l p{3.3cm} l}
\toprule
Index & Reads as & Formula & Standard equivalent & Clinical origin \\
\midrule
L & confidence kept on errors & $b/(b+d)$ & false-positive rate of KEEP & MMPI-3 L; PAI PIM \\
Fp & confidence withdrawn on correct items & $c/(a+c)$ & false-negative rate of KEEP & MMPI-3 Fp; PAI MAL \\
RBS & directional contrast & $\text{Fp} - (1-\text{L})$ & $-J$; binary AUROC $= (1-\text{RBS})/2$ & MMPI-3 RBS \\
TRIN & response dominance & $\max(n_K,n_W)/N$ & modal-response share & MMPI-3 TRIN \\
$r$ & item sensitivity & Pearson on 0/1 vectors & $\phi$ coefficient & -- \\
\bottomrule
\end{tabular}
\end{table}

L is the share of errors on which the model kept its answer: $L = .95$ means confidence was retained on 19 of 20 errors. Fp is the share of correct answers the model withdrew. RBS is the withdrawal rate on correct items minus the withdrawal rate on incorrect items. Because $\text{RBS} = -J$, the statements RBS $> 0$, binary AUROC $< .5$ and $r < 0$ are equivalent: confidence points the wrong way. Two indices from the original derivation, K (the share of errors on which the model bets) and F (withdrawal on items that at least 85\% of models keep), need a second probe or a multi-model norm and are used only descriptively for the battery.

Separating L and Fp is useful because an uninformative signal can arise in different ways. A binary AUROC of .5 can come from keeping everything ($L = 1$, $\text{Fp} = 0$), withdrawing everything ($L = 0$, $\text{Fp} = 1$) or responding at random; perfect inversion gives an AUROC of 0. These have different causes and remedies: re-eliciting with a probe that permits low confidence, choosing a different threshold, or not using the signal. The aggregate statistic does not distinguish them.

\subsection{Ordered sequence, thresholds and tiers}

\begin{figure}[t]
\centering
\includegraphics[width=0.72\textwidth]{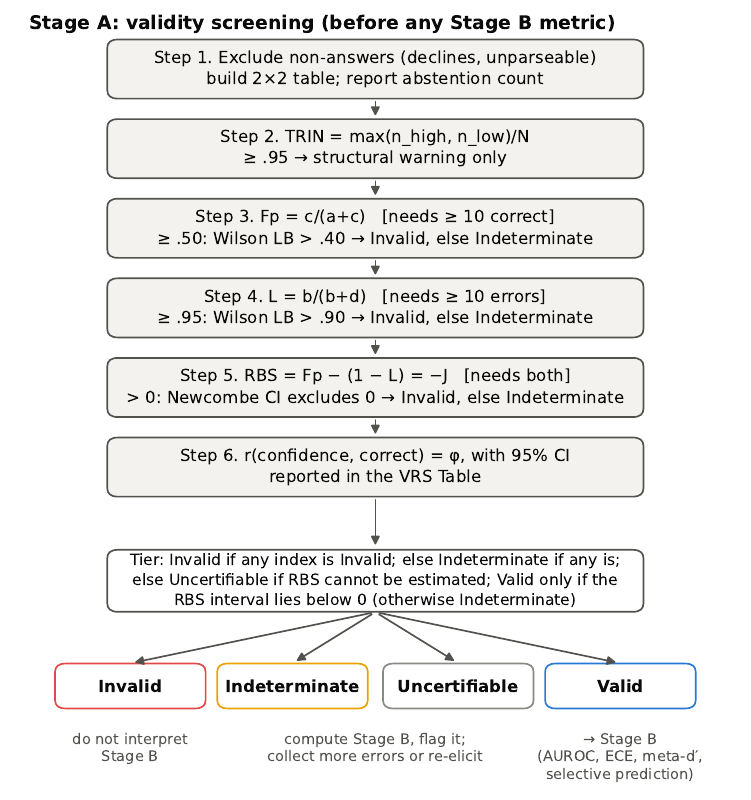}
\caption{The screening sequence. Each index is computed only if its denominator reaches 10. Indeterminate means either that a threshold is exceeded on the point estimate without meeting the interval guard, or that no flag fires but item sensitivity cannot be detected.}
\label{fig:protocol}
\end{figure}

Figure~\ref{fig:protocol} gives the sequence. Each index is computed only when its own denominator reaches 10 (correct items for Fp, errors for L, both for RBS).
\begin{enumerate}
\item Exclude non-answers and pipeline-supplied values; build the table; report the exclusions.
\item TRIN. If $\text{TRIN} \geq .95$, record a structural warning. TRIN never decides a tier, because high accuracy with moderate L produces high TRIN mechanically.
\item Fp. If $\text{Fp} \geq .50$: Invalid if the Wilson lower bound \citep{wilson1927} exceeds .40, otherwise Indeterminate.
\item L. If $L \geq .95$: Invalid if the Wilson lower bound exceeds .90, otherwise Indeterminate.
\item RBS, with the Newcombe hybrid score interval \citep{newcombe1998}. If $\text{RBS} > 0$: Invalid if the interval excludes 0, otherwise Indeterminate.
\item $r$ with 95\% interval, always reported.
\end{enumerate}
The tier is Invalid if any index is Invalid; otherwise Indeterminate if any index is Indeterminate; otherwise Uncertifiable if RBS cannot be estimated. A model reaching this point is \emph{Valid only if the RBS interval lies entirely below zero}, that is, if $J$ is significantly positive; otherwise it is Indeterminate (no detectable item sensitivity). This last condition is new in version 0.3. Without it, Valid meant only that no response set had been detected, and an uninformative random policy was classified Valid in up to half of simulated draws (Section~\ref{sec:synth}). The condition plays the role that the inconsistency scales play in clinical instruments. The MMPI-3 does not treat a profile as interpretable merely because its over- and under-reporting scales are quiet; it also checks VRIN-r and TRIN-r for random or content-unrelated responding \citep{benporath2020mmpi3}, and the PAI does the same with ICN and INF \citep{morey2007}. L, Fp and RBS detect fixed and inverted response sets but not random responding. Requiring that confidence measurably track correctness supplies that missing check.

The thresholds are boundaries, not tuned operating points. $L \geq .95$ means at most one error in 20 is withdrawn, too few for an abstention rule to act on. $\text{Fp} \geq .50$ means most correct answers are withdrawn. $\text{RBS} > 0$ is the minimal condition for inversion. The interval guards stop a point estimate that sits just past a threshold from being treated as established.

\paragraph{Sample-size requirement.} For the Wilson lower bound on L to exceed .90, a model needs at least 35 errors if it keeps every one ($L = 1.00$), 71 at $L = .97$, 97 at $L = .96$, and 139 at $L = .95$ (treating the observed L as fixed). High-accuracy models seldom make that many errors on a few hundred items: 10 of the 20 battery models made fewer than 35. For such models the screen can report that L is high but cannot certify blanket confidence; this is what the Indeterminate tier records. Item counts should be planned from the expected number of errors, not the number of items.

\subsection{Continuous confidence: the cut point}\label{sec:binarisation}

Verbalized 0--100 confidence must be binarised before the table can be built: confidence $\geq c$ counts as high. The cut point changes the question. At cut point $c$ the screen asks whether confidence is usable \emph{as a threshold at $c$}. If almost every answer sits above $c$, L approaches 1 and the screen reports blanket confidence; that is a correct description of the signal at that operating point, not an error. A per-model median split, which an earlier draft of this version used, is not suitable for screening. With confidence piled on a few values, its verdict is decided by whether ties at the median count as high or low (sending ties low turned 31 of 33 Valid atlas models into 28 of 33 Invalid), and an exact 50/50 split places Fp near the .50 threshold by construction. Whether a model's confidence \emph{ranks} its answers well is a Stage B question, answered by the AUROC on the raw confidence. We therefore report verdicts at a range of fixed cut points, state the cut point with every verdict, and recommend screening at the threshold a deployment will use.

\subsection{Reporting}

For each model we propose a VRS (Validity Report for confidence Screening) table: model and version; benchmark; items, exclusions, correct and incorrect counts; elicitation method and original confidence format; cut point; TRIN; L, Fp and RBS with intervals; $r$ with interval and $p$; tier; and the reason for any flag. Table~\ref{tab:battery} is the VRS table for the battery in compact form.

\section{Data and Analysis}\label{sec:data}

\paragraph{Battery (binary probes).} Twenty models from seven families answered a 524-item battery in six tracks \citep{cacioli2026c}: Claude Opus 4.6, Sonnet 4.6 and Haiku 4.5; GPT-5.4, 5.4 mini and 5.4 nano; Gemini 2.5 Flash, 2.5 Pro, 3 Flash and 3.1 Pro; Gemma 3 1B, 12B and 27B; Qwen3 235B, Next 80B Instruct, Next 80B Thinking and Coder 480B; DeepSeek V3.2 and R1; and GLM-5. After each answer the model was asked whether to KEEP or WITHDRAW it and whether to BET that it was correct. The screen uses the KEEP/WITHDRAW decision.

Five tracks (attention, executive function, learning, metacognition and social cognition; 452 items) are two-alternative items scored against a key. The sixth (prospective regulation; 72 items) asks for free-text answers and was graded by string matching. That grading is unreliable in both directions. It marks correct answers wrong (``H\textsubscript{2}O'' for the key ``H2O'', ``ln$|x|$ + C'' for ``ln(x)'', ``(1+$\sqrt{5}$)/2'' for ``golden ratio''): under narrow notation-only equivalence rules, 83 of the track's 168 scored errors are correct answers, 80 of them KEEPs (an independent, stricter regrade found 73). It marks some wrong answers correct (``2.7141'' accepted for the key 2.7183; a ``Yes'' accepted for a ``No'' key), and several keys are disputable (the Hubble constant, the longest isogram). It is also where R1 declined: 71 of R1's 72 responses in this track are declines. \emph{The primary analysis therefore uses the five keyed tracks (452 items, 9,040 responses).} All six tracks with non-answers excluded are reported as a sensitivity analysis. In the five keyed tracks the release contains the key but not the model's own response text, so non-answers and grading there cannot be audited; we return to this in the limitations.

\paragraph{Atlas MMLU (verbalized confidence).} Thirty-three models from eight families each answered up to 1,500 MMLU items, stratified over six subject groups, and reported a confidence from 0 to 100 \citep{cacioli2026atlas}; 12 runs are partial (598 to 1,499 items). For GPT-oss-120B the parser substituted a default of 50 when no confidence could be extracted; 980 of its 1,500 values are this default, and the parsing log records no genuine 50 for that model. Those rows are excluded, along with 13 unparsed answers across all models, leaving 46,161 scored responses. Accuracy is similar on GPT-oss-120B's imputed and retained rows (.900 and .892), but its 520 retained rows are not a random sample, and its results are conditional on that exclusion.

\paragraph{MMLU-500.} The 20 battery models answered a separate sample of 500 MMLU items with the same elicitation (8,568 scored responses; seven runs are partial, 114 to 498 items). Only 71 of the 500 questions are also in the atlas sample.

\paragraph{Yang et al.\ (2024).} \citet{yang2024} released item-level predictions for 11 models (Gemma 1.1 2B and 7B, GPT-3.5-turbo, GPT-4o, GPT-4o-mini, Llama 3 8B and 70B, Qwen1.5 7B, 32B, 72B and 110B) on 10 benchmarks (ARC-Challenge, ARC-Easy, CommonsenseQA, LogiQA, MMLU, SciQ, Social IQa, TriviaQA, TruthfulQA MC1 and MC2) under 17 prompting methods, with confidence on a 0--1 scale. Our primary analysis uses their basic single-sample verbalized prompt (110 model--benchmark cells, 112,909 responses); the other 16 methods are a sensitivity analysis. For TriviaQA and TruthfulQA MC2 the released correctness is fractional partial credit (22 of the 110 cells); we count a score of at least 0.5 as correct, and counting any credit as correct leaves the verdicts at the 0.9 cut point unchanged.

\paragraph{Analysis.} All verdicts come from \texttt{validity\_screen} 0.3.0 with the constants above and no manual adjustment; the version 0.2 verdict (without the item-sensitivity condition) is reported alongside. Continuous AUROC intervals are 2,000-draw bootstrap percentiles. Split-half stability draws 1,000 random halves of each model's items, screens one half per draw, and records how often it reproduces the full-sample tier. Selective accuracy averages over ties in confidence. A single script regenerates every number, table and figure in this paper from the released data (see Data and code).

\paragraph{What was decided after seeing the data.} The thresholds and interval guards come from version 1 and are unchanged. Everything else that shapes the verdicts reported here was settled on these same data: the non-answer rule and denominator guards (version 0.2), the item-sensitivity condition for Valid (version 0.3), the exclusion of the free-text track, and the cut points used for verbalized confidence. Each change is listed with its reason in Appendix~\ref{app:amend}, and each was prompted by an error or a failure found in review rather than by the verdicts it produces, but none has been tested on held-out binary-probe data. The verdicts below are therefore descriptions of these models under this protocol, not validation of the protocol.

\section{Results}

\subsection{Policies with known properties (RQ1)}\label{sec:synth}

Eleven response policies were simulated, each specified by its probability of KEEP on correct and on incorrect items, with KEEP drawn independently per item over 2,000 draws (Table~\ref{tab:synthetic}). The item set has 500 items and either 40 errors (92\% accuracy, the battery median) or 150 errors. Deterministic patterns are always classified as intended: always KEEP, always WITHDRAW and inversion are Invalid, and good monitors are Valid. Three results describe behaviour at the boundaries. A weak monitor (keeping 90\% of correct and 75\% of incorrect answers) is Valid in 77\% of draws at 40 errors and 99\% at 150. Uninformative random policies are Valid in only 2--4\% of draws under version 0.3, against about half for a random 80\% keeper under version 0.2, which is why the item-sensitivity condition was added. Near-blanket confidence (keeping 97\% of everything) is certified Invalid in 30\% of draws at 40 errors and 91\% at 150; with few errors it is usually Indeterminate.

\begin{table}[t]
\centering
\caption{Simulated policies: percentage of 2,000 draws classified Valid (V), Indeterminate (I) or Invalid (X) by version 0.3, with 500 items and 40 or 150 errors. KEEP\% gives P(KEEP $\mid$ correct) / P(KEEP $\mid$ incorrect). No draw was Uncertifiable.}
\label{tab:synthetic}
\small
\begin{tabular}{lrrrrrrr}
\toprule
 & & \multicolumn{3}{c}{40 errors} & \multicolumn{3}{c}{150 errors} \\
\cmidrule(lr){3-5}\cmidrule(lr){6-8}
Policy & KEEP\% & V & I & X & V & I & X \\
\midrule
Always KEEP & 100 / 100 & 0 & 0 & 100 & 0 & 0 & 100 \\
Always WITHDRAW & 0 / 0 & 0 & 0 & 100 & 0 & 0 & 100 \\
Random, 50\% KEEP & 50 / 50 & 2 & 46 & 51 & 3 & 45 & 52 \\
Random, 80\% KEEP & 80 / 80 & 3 & 95 & 2 & 2 & 95 & 3 \\
Random, 95\% KEEP & 95 / 95 & 4 & 84 & 12 & 3 & 46 & 51 \\
Near-blanket confidence (97/97) & 97 / 97 & 4 & 66 & 30 & 3 & 6 & 91 \\
Perfect monitor & 100 / 0 & 100 & 0 & 0 & 100 & 0 & 0 \\
Noisy monitor (80/40) & 80 / 40 & 100 & 0 & 0 & 100 & 0 & 0 \\
Weak monitor (90/75) & 90 / 75 & 77 & 23 & 0 & 99 & 1 & 0 \\
Weakly inverted (85/95) & 85 / 95 & 0 & 60 & 40 & 0 & 8 & 92 \\
Inverted (20/80) & 20 / 80 & 0 & 0 & 100 & 0 & 0 & 100 \\
\bottomrule
\end{tabular}
\end{table}

\subsection{Battery classification (RQ2)}

\begin{table}[p]
\centering
\caption{VRS summary for the battery (five keyed tracks, 452 items per model, binary KEEP/WITHDRAW probe). Err.: errors. Brackets: 95\% Wilson intervals (L, Fp) and Newcombe intervals (RBS). --, n.e.: not estimable (fewer than 10 errors). In parentheses: the version 0.2 tier where it differs. Sorted by tier, then $r$.}
\label{tab:battery}
\resizebox{\textwidth}{!}{%
\begin{tabular}{lrrllllrl}
\toprule
Model & Err. & Acc. & L & Fp & RBS & $r$ & TRIN & Tier \\
\midrule
DeepSeek-R1 & 6 & .987 & -- & .946\,[.921, .964] & n.e. & +.027 & .947 & Invalid \\
GPT-5.4 nano & 87 & .808 & .517\,[.414, .619] & .378\,[.330, .429] & $-$.105\,[$-$.219, +.009] & +.084 & .602 & Indeterminate (Valid) \\
Gemini 3 Flash & 22 & .951 & .909\,[.722, .975] & .028\,[.016, .048] & $-$.063\,[$-$.251, +.006] & +.078 & .969 & Indeterminate (Valid) \\
Qwen 80B Think & 27 & .940 & .963\,[.817, .993] & .007\,[.002, .021] & $-$.030\,[$-$.176, +.003] & +.076 & .991 & Indeterminate \\
Gemini 3.1 Pro & 24 & .947 & .958\,[.798, .993] & .040\,[.025, .063] & $-$.002\,[$-$.163, +.039] & +.002 & .960 & Indeterminate \\
GLM-5 & 20 & .956 & .800\,[.584, .919] & .007\,[.002, .020] & $-$.193\,[$-$.409, $-$.073] & +.322 & .985 & Valid \\
Gemini 2.5 Flash & 22 & .951 & .864\,[.667, .953] & .002\,[.000, .013] & $-$.134\,[$-$.331, $-$.045] & +.308 & .991 & Valid \\
Gemini 2.5 Pro & 30 & .934 & .900\,[.744, .965] & .000\,[.000, .009] & $-$.100\,[$-$.256, $-$.034] & +.307 & .993 & Valid \\
Claude Haiku 4.5 & 47 & .896 & .404\,[.276, .547] & .193\,[.157, .234] & $-$.403\,[$-$.536, $-$.255] & +.290 & .765 & Valid \\
Sonnet 4.6 & 32 & .929 & .719\,[.546, .844] & .033\,[.020, .055] & $-$.248\,[$-$.421, $-$.120] & +.289 & .949 & Valid \\
Qwen 80B Inst & 45 & .900 & .733\,[.590, .840] & .052\,[.034, .078] & $-$.215\,[$-$.360, $-$.105] & +.248 & .927 & Valid \\
Qwen 235B & 42 & .907 & .667\,[.516, .790] & .088\,[.064, .119] & $-$.246\,[$-$.399, $-$.118] & +.227 & .889 & Valid \\
Gemma 3 12B & 65 & .856 & .446\,[.332, .567] & .289\,[.246, .336] & $-$.264\,[$-$.387, $-$.135] & +.198 & .673 & Valid \\
Qwen Coder 480B & 40 & .912 & .575\,[.422, .715] & .158\,[.126, .196] & $-$.267\,[$-$.424, $-$.122] & +.197 & .819 & Valid \\
GPT-5.4 & 28 & .938 & .607\,[.424, .764] & .125\,[.097, .160] & $-$.268\,[$-$.453, $-$.107] & +.185 & .858 & Valid \\
Opus 4.6 & 28 & .938 & .857\,[.685, .943] & .026\,[.015, .046] & $-$.117\,[$-$.289, $-$.029] & +.157 & .967 & Valid \\
GPT-5.4 mini & 49 & .892 & .694\,[.555, .805] & .134\,[.104, .171] & $-$.172\,[$-$.314, $-$.055] & +.149 & .847 & Valid \\
Gemma 3 27B & 56 & .876 & .893\,[.785, .950] & .028\,[.016, .049] & $-$.079\,[$-$.188, $-$.018] & +.137 & .962 & Valid \\
DeepSeek V3.2 & 51 & .887 & .824\,[.697, .904] & .067\,[.047, .096] & $-$.109\,[$-$.237, $-$.023] & +.128 & .920 & Valid \\
Gemma 3 1B & 138 & .695 & .594\,[.511, .673] & .299\,[.251, .352] & $-$.106\,[$-$.203, $-$.012] & +.104 & .668 & Valid \\
\bottomrule
\end{tabular}}
\end{table}

The screen classifies one model as Invalid, four as Indeterminate and 15 as Valid (Table~\ref{tab:battery}).

\textbf{DeepSeek-R1 (Invalid: blanket withdrawal).} R1 withdrew 95\% of its correct answers ($\text{Fp} = .946$, $[.921, .964]$) and kept only 24 of 452 answers. It made 6 errors, all in the metacognition track, so L and RBS cannot be estimated. It withdrew all 6, and all 24 answers it kept were correct; with six errors nothing can be said about whether its withdrawals track correctness. The Invalid verdict therefore rests on Fp alone and says that the KEEP signal is unusable as a filter, since it discards 95\% of correct answers, not that it is inverted. Its second probe points the other way: R1 bet on 5 of its 6 errors but on only 39\% of its correct answers (Fisher $p = .036$), and its ordinal AUROC combining both probes is .319 (bootstrap interval $[.22, .50]$). It chose WITHDRAW together with BET on 37\% of its correct answers; the next highest rates are Gemma 3 1B (17\%) and GPT-5.4 nano (3\%). Version 1 described R1's KEEP/WITHDRAW signal as catastrophically inverted; that pattern came from 71 declines in the free-text track, which the harness had logged as KEEP+BET and the grader scored as errors.

\begin{figure}[t]
\centering
\includegraphics[width=\textwidth]{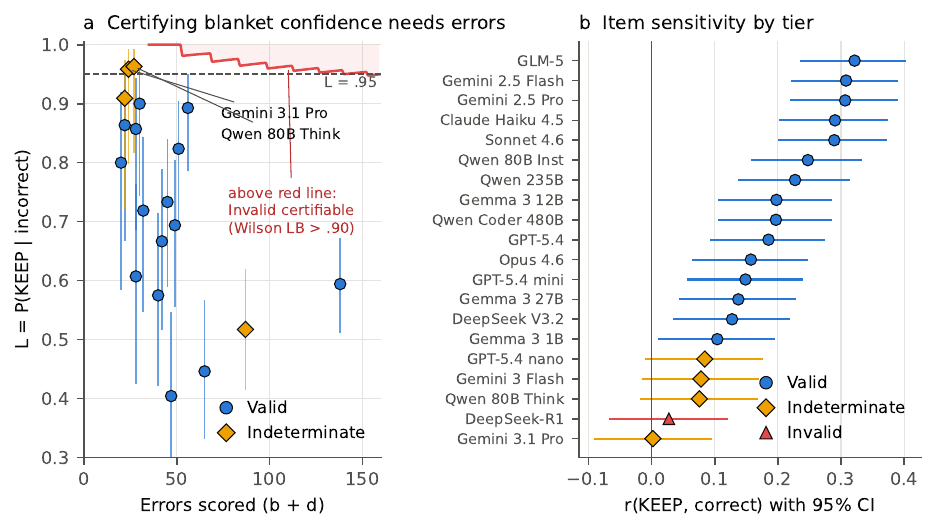}
\caption{Battery (five keyed tracks). (a) L against the number of errors, with Wilson intervals. The red line is the smallest L whose Wilson lower bound exceeds .90 at that error count; blanket confidence can be certified only above it. The two high-L Indeterminate models sit above the .95 threshold but below the line. DeepSeek-R1 is omitted (6 errors; L not estimable). (b) Item sensitivity $r$ with 95\% intervals.}
\label{fig:battery}
\end{figure}

\textbf{Gemini 3.1 Pro and Qwen3 Next 80B Thinking (Indeterminate: high L, too few errors).} Gemini 3.1 Pro kept its answer on 23 of its 24 errors ($L = .958$, lower bound .798), and its withdrawals were unrelated to correctness ($r = .002$). Qwen 80B Thinking kept 26 of 27 ($L = .963$, lower bound .817; $r = .076$). Both exceed the L threshold, but neither made enough errors to certify it (Figure~\ref{fig:battery}a). The same base model without reasoning mode, Qwen 80B Instruct, has $L = .733$ and $r = .248$.

\textbf{Gemini 3 Flash and GPT-5.4 nano (Indeterminate: no detectable item sensitivity).} Neither triggers a threshold, but neither shows withdrawals that measurably track correctness: Gemini 3 Flash has $r = .078$ (RBS interval $[-.251, +.006]$) and GPT-5.4 nano $r = .084$ ($[-.219, +.009]$). Under version 0.2 both were Valid.

\textbf{Valid models.} The 15 Valid models have $r$ between .104 and .322 and L between .40 and .90. Eight models have $\text{TRIN} \geq .95$, five of them Valid, which is why TRIN is a warning only. No battery model has an RBS interval above zero, so none shows confidence that points the wrong way on the KEEP probe.

\paragraph{What the screen adds to a direct test.} The five models that are not Valid are exactly the five whose $r$ is not significantly positive. That follows from the version 0.3 rule, under which Valid requires the item-sensitivity test to pass, so the screen is never less sensitive than that test. It is also no more sensitive, and comparing $r$ between the Valid and non-Valid groups would only restate how the groups were formed. What the screen adds is the reason for each failure. R1 withdraws almost everything. Gemini 3.1 Pro and Qwen Thinking keep almost everything, with too few errors to certify it. Gemini 3 Flash and GPT-5.4 nano show no response set and no detectable signal. These call for different actions: re-eliciting, collecting more errors, or not using the signal.

\paragraph{Sensitivity.} With all six tracks (non-answers excluded) the counts are the same (15 / 4 / 1), but Gemma 3 1B replaces GPT-5.4 nano among the Indeterminate models; both sit at the edge of detectable sensitivity. Moving the L threshold anywhere from .93 to .97 changes no tier.

\subsection{Stability (RQ3)}

Verdicts need errors, and halving the items halves the errors. In 1,000 random half-samples (226 items), the half reproduced the full-sample tier at least 90\% of the time for 6 of the 20 models. Agreement was lowest for Gemma 3 1B (.27), DeepSeek V3.2 (.49), Gemma 3 27B (.53), Gemini 3 Flash (.54) and GLM-5 (.58), whose RBS intervals approach zero or whose error counts fall below what certification needs. Instability is therefore predictable from error counts and interval widths. A single administration of a few hundred items gives a firm verdict only for models far from every boundary.

\subsection{Selective prediction (descriptive)}

Version 1 and its companion presented the screen as predicting selective-prediction performance. That comparison is close to circular. Type 2 AUROC was computed on the same items from the same KEEP decision plus one BET decision, and for a binary signal RBS and AUROC are the same quantity. Across models RBS correlates $-.73$ with the ordinal AUROC ($n = 19$; R1's RBS is not estimable). We report it only as description. Ordinal AUROC averages .644 for Valid models, .532 for Indeterminate ones and .319 for R1. Because the ordinal signal has four levels, selective accuracy moves in a single step for almost every model. Sonnet 4.6 rises from 92.9\% at full coverage to 96.9\% at 50\%. Gemini 3.1 Pro (94.7\% to 95.0\%) and Qwen 80B Thinking (94.0\% to 94.2\%) barely move, and R1 stays near 98\% throughout. Version 1 reported that R1's accuracy fell to 11\% at 10\% coverage; that came from the scored declines.

\subsection{Reasoning variants and families (descriptive)}

Two pairs share a base model or lineage. From DeepSeek V3.2 to R1, withdrawal of correct answers rises from 6.7\% to 94.6\% and K from .275 to .833. From Qwen 80B Instruct to Thinking, L rises from .733 to .963 and the withdrawal difference between incorrect and correct items falls from .215 to .030. The reasoning variants move in opposite directions, one towards withdrawing and one towards keeping. These are two pairs, the DeepSeek models differ in more than reasoning training, and Qwen Thinking's L cannot be certified, so this is an observation. The four Gemini models share high L (mean .908, range .864 to .958), which is why we recommend comparing models within a family as well as against cross-family thresholds; in clinical assessment, norms calibrated on one population misclassify another \citep{cheung1996,vandevijver2005}. Within GPT-5.4, L is not monotone in model size (nano .517, mini .694, full .607).

\subsection{Verbalized confidence (RQ4)}\label{sec:cross}

\begin{figure}[t]
\centering
\includegraphics[width=\textwidth]{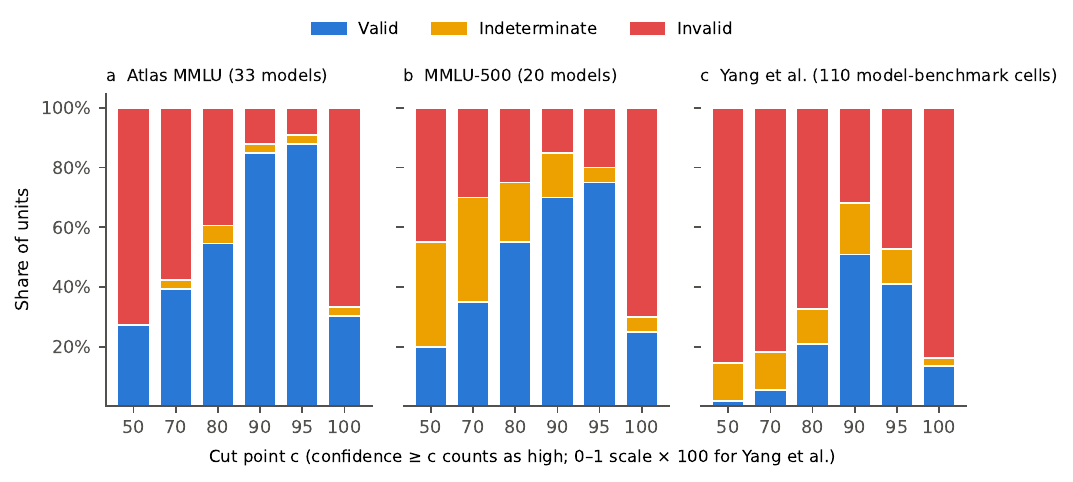}
\caption{Tier composition at six cut points for verbalized confidence. Units are models (a, b) or model--benchmark cells (c; Yang et al.'s \texttt{basic\_1s} prompt, confidence $\times 100$).}
\label{fig:cuts}
\end{figure}

\paragraph{The verdict depends on the cut point.} Verbalized confidence piles at the top of the scale: the median model gave 90 or more on 91\% of atlas items and 93\% of MMLU-500 items. Figure~\ref{fig:cuts} shows verdicts at six cut points. At 50 almost nothing is low-confidence, and the screen reports blanket confidence: 24 of 33 atlas models, 9 of 20 MMLU-500 models (with 7 more Indeterminate) and 94 of 110 Yang cells are Invalid. Between 90 and 95, where models actually divide their answers, most are Valid: 28 to 29 of 33 atlas models, 14 to 15 of 20 MMLU-500 models, and about half of the Yang cells. At 100 the verdict reverses again, because most correct answers fall below the cut. Sending ties at 50 the other way ($> 50$) changes one atlas model and two MMLU-500 models. For most models, then, stated confidence is unusable as an abstention threshold at the midpoint of its own scale. A Valid verdict near the top means only that the cut separates errors from correct answers better than chance, not that it separates them well: Gemma 3 12B is Valid at 95 in both MMLU runs with a raw-confidence AUROC of .505 (interval $[.473, .534]$ in the atlas). Reading verdicts across six cut points also raises the chance that a null signal passes at one of them, so a screen should use one cut point chosen in advance, ideally the deployment threshold.

\paragraph{Ranking.} Ranking quality is a Stage B property, separate from the screen. AUROC on the raw confidence ranges from .498 to .806 in the atlas (mean .697) and from .500 to .837 in MMLU-500. Two atlas models are never Valid at any cut point. One is Gemma 3 1B, which gave exactly 100 on 96\% of items (AUROC .498, interval $[.487, .508]$). The other is Gemma 3 4B, whose signal is real but very weak (AUROC .528, $[.502, .553]$). Gemma 3 12B (.505) and 27B (.566) are Invalid at both 50 and 90. In MMLU-500 the models never Valid are Gemma 3 1B and the two shortest runs (DeepSeek V3.2, 114 items; Qwen Coder 480B, 154 items), whose AUROC intervals include .5.

\paragraph{Replication across MMLU samples.} For the 20 models in both MMLU runs, raw-confidence AUROC correlates $r = .88$ between the two item samples. Tier agreement rises with the cut point: 65\% at 50, 80\% at 90 and 90\% at 95. Near the bottom of the scale, a verdict depends on how many of a model's few low-confidence answers happen to fall on errors, and that varies with the items sampled.

\paragraph{Yang et al.\ (2024).}\label{sec:yang} Qwen1.5-7B-chat is Invalid on all 10 benchmarks at both 50 and 90. Its confidence barely varies (mean within-cell SD .011 on the 0--1 scale; at most six distinct values), and its AUROC averages .501 and never exceeds .508. Gemma 1.1 7B is also Invalid on all 10 benchmarks at 90, with L between .97 and 1.00, certified in every cell; its AUROC averages .520. GPT-3.5-turbo is Invalid on 7 of 10. AUROC across all cells averages .607 (.43 to .89). Prompting matters as much as the model: across Yang et al.'s 17 methods, the share of cells Invalid at 90 ranges from 19\% to 60\%. Version 1 reported one Invalid model, a mean AUROC of .624, a correlation of $+.894$ between L and AUROC, and model and benchmark lists that do not match this data set. Those results were not computed from the released data and are withdrawn.

\begin{figure}[t]
\centering
\includegraphics[width=0.56\textwidth]{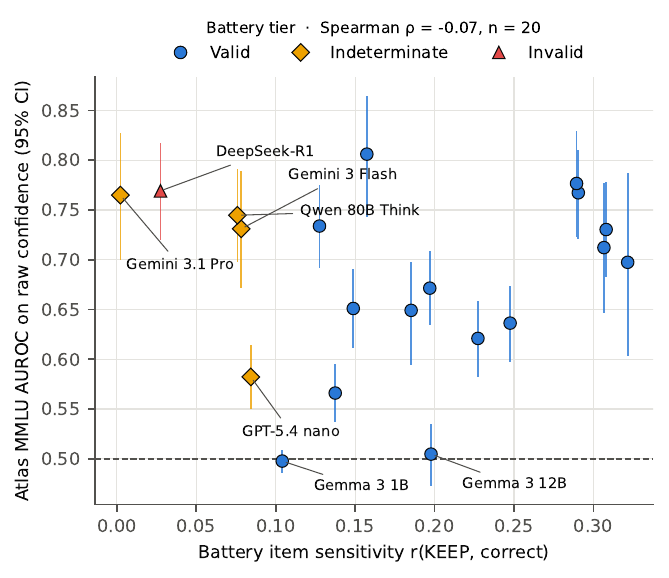}
\caption{Battery item sensitivity (binary probe) against atlas MMLU AUROC on raw verbalized confidence (95\% bootstrap intervals) for the 20 models in both. Markers show the battery tier. The binary-probe result does not predict discrimination under verbalized confidence.}
\label{fig:transfer}
\end{figure}

\paragraph{Verdicts do not carry over between probe formats.} The three battery models whose binary signal failed or could not be certified (R1, Gemini 3.1 Pro, Qwen 80B Thinking) discriminate well under verbalized confidence: atlas AUROC .769, .765 and .745, and Valid at 90 in both MMLU runs (Figure~\ref{fig:transfer}). In the other direction, Gemma 3 1B, 12B and 27B are Valid on the battery but have MMLU AUROC between .50 and .58 and are Invalid at 50 and at 90 in both MMLU runs (12B and 27B pass at 95, by the weak criterion just described). Across the 20 models, battery item sensitivity is unrelated to MMLU AUROC (Spearman $\rho = -.07$ with the atlas, $+.07$ with MMLU-500). Version 1 stated that a model passing the binary screen would also pass under continuous confidence; the Gemma 3 models contradict that at the cut points 50 and 90. A verdict describes a model under a particular probe, task and cut point, and has to be obtained under the conditions it is meant to support.

\section{Discussion}

\paragraph{What the screen is for.} The screen answers one narrow question before any calibration, metacognition or selective-prediction metric is reported: is this signal dominated by a response set, which one, and is there evidence that it tracks correctness at all? On binary probes it found blanket withdrawal in one model, near-blanket confidence that could not be certified in two, and no detectable sensitivity in two more. On verbalized confidence it showed something an AUROC does not: for most models, stated confidence at the midpoint of its scale is effectively a constant, even when its ranking is informative.

\paragraph{What it does not do.} It is not a more sensitive detector than a significance test on $r$, which it now contains. It cannot certify blanket confidence for a model that makes few errors. Its verdicts are not stable at a few hundred items for models near any boundary. It is not a criterion-validated predictor of deployment performance; the version 1 claim to that effect rested on a circular comparison and on a scoring artefact.

\paragraph{The corrections themselves.} Three of the errors corrected here are of one kind: a value the pipeline supplied was read as a value the model produced. The battery's declines were logged as KEEP+BET, the string-match grader's verdicts stood in for correctness, and the atlas parser's defaults stood in for confidence. Each manufactured or concealed a response set. Step 1 of the screen exists for this reason, and it is the step most easily skipped.

\paragraph{Recommendations.}
\begin{enumerate}
\item Exclude non-answers, parser defaults and other pipeline-supplied values before scoring; report the counts; release raw responses so others can check.
\item Grade free-text answers with normalised matching and manual adjudication, not substring tests.
\item Plan item counts from expected errors: certifying blanket confidence needs 35 errors at best and about 140 near the threshold.
\item For continuous confidence, choose one cut point in advance (the deployment threshold if there is one), screen at it and state it; report ranking separately, as AUROC on the raw values.
\item Screen under the probe format, task and prompt the signal will be used with; verdicts did not carry over between them.
\item Report $r$ and its interval with every tier.
\end{enumerate}

\paragraph{Limitations.} The battery has 20 models from seven families and five non-Valid verdicts. In its five keyed tracks the release holds the key but not the model's response text, so non-answers and grading there cannot be audited, and harness defaults like those found in the free-text track cannot be ruled out. Verdicts are single administrations: the two MMLU samples replicate the verbalized-confidence results, but nothing replicates the binary battery, and the protocol changes were settled on the same battery they are applied to. Thresholds are logical boundaries, not values derived from deployment costs. The indices, $r$ and AUROC are computed from the same responses, so agreement among them is partly algebraic. The causes of the response sets (training, reasoning mode or elicitation) are not identified.

\paragraph{Status of related papers.}\label{sec:upstream} The battery paper \citep{cacioli2026c} uses the same response files, including the free-text track's grading and declines; its results for that track, and any result that includes it, should be read in light of Section~\ref{sec:data}. The atlas paper \citep{cacioli2026atlas} uses the MMLU data analysed here. Its GPT-oss-120B results include the 980 imputed confidences, and its description of three battery models as Invalid and of DeepSeek-R1 as inverted relies on version 1 of this paper. Both will be revised.

\section{Conclusion}

Clinical assessment checks whether a response pattern is interpretable before interpreting it. For LLM confidence the check is cheap: one table, three rates and their intervals. In this corrected analysis it found blanket withdrawal, uncertifiable near-blanket confidence and undetectable sensitivity on binary probes. It also showed that verbalized confidence is mostly a ceiling at the midpoint of its scale but often an informative ranking. The broader lesson came from the corrections: every dramatic result in version 1 came from values the pipeline had supplied rather than values the models produced, and an ordered check of the inputs would have caught them.

\section*{Data, code and use of AI}\sloppy
Data, analysis code (including \texttt{run\_all.sh}, which regenerates every number here) and output files are at \url{https://github.com/synthiumjp/validity-scaling-llm}; the atlas data are at \url{https://github.com/synthiumjp/metacognitive-profile-atlas}. Yang et al.'s predictions are available from the original authors. The screening tool is \texttt{validity\_screen} 0.3.0. The author received no external funding and declares no competing interests. Claude (Anthropic) was used for code generation, reanalysis and manuscript preparation, and for two independent reviews that found the errors corrected here. All scientific decisions were made by the author.

\bibliographystyle{plainnat}
\bibliography{references}

\appendix

\section{Protocol amendments}\label{app:amend}

\begin{center}
\small
\begin{tabular}{p{4.2cm} p{4.6cm} p{5.6cm}}
\toprule
Version 0.1 (v1 papers) & Version 0.3 (this paper) & Reason \\
\midrule
Non-answers scored with harness defaults & Excluded before scoring, with pipeline-supplied values; counts reported & A declined item carries no confidence judgement about an answer \\
Any cell $< 5$ $\Rightarrow$ ``insufficient data'' & Each index computed when its own denominator $\geq 10$; Uncertifiable tier & Blanket confidence empties cell $d$ by definition, so the old guard blocked the response sets the screen targets; it returned no tier for seven battery models \\
Wald interval for RBS & Newcombe hybrid score interval & Wald intervals collapse at $L = 1$ or $\text{Fp} = 0$ \\
RBS Invalid also required point estimate $> .05$ (v1 text) & RBS $> 0$ with interval excluding 0 & One rule, matching the implementation \\
Valid = no flag & Valid = no flag \emph{and} RBS interval below 0 & An uninformative random policy was Valid in up to half of draws (Section~\ref{sec:synth}); the screen had no counterpart to the MMPI-3 VRIN-r or PAI INF check for random responding \\
Continuous confidence binarised at 50 or the median & Fixed cut points, stated with each verdict; no median split & The median verdict depends on tie handling; an exact median split puts Fp near .50 by construction \\
Tiers adjustable in analysis code & Tier is a function of data and constants only & Reproducibility \\
\bottomrule
\end{tabular}
\end{center}
\noindent The thresholds (Fp $\geq .50$, L $\geq .95$, RBS $> 0$) and interval guards (.40, .90, 0) are unchanged.

\section{Corrections to version 1 and its companions}\label{app:corrections}

\begin{center}
\small
\begin{tabular}{p{6.6cm} p{7.8cm}}
\toprule
Version 1 claim & Corrected \\
\midrule
DeepSeek-R1 shows inverted monitoring ($r = -.798$, RBS $= +.868$) & Artefact of 71 declines in the free-text track scored as errors (139 of all 142 non-answers carried KEEP+BET). R1 withdraws 95\% of correct answers; KEEP $r = +.027$; L and RBS not estimable (6 errors). Its BET probe does favour errors (ordinal AUROC .319) \\
R1 KEEP$\times$BET table (17707): 77 / 0 / 171 / 0 & Five keyed tracks: KEEP+BET 6, KEEP+NO BET 18, WITHDRAW+BET 171, WITHDRAW+NO BET 257 \\
R1 selective accuracy falls from 85\% to 11\% at 10\% coverage & Artefact; R1 stays near 98\% \\
Tiers: 3--4 Invalid, 2--3 Indeterminate, 14 Valid & 1 Invalid, 4 Indeterminate, 15 Valid (five keyed tracks, v0.3) \\
Group difference in $r$, $d = 2.41$--$2.48$, $p = .001$; robust to leave-one-out & Under v0.3 the groups are defined by the significance of $r$, so the comparison is not reported \\
Screen predicts selective prediction ($d = 2.81$, $\eta^2 = .47$); monotone tier ordering; split-half $d = 1.77$ & Criterion computed from the same items and probe; reported descriptively. Split-half used fixed labels; replaced by tier reproduction \\
Classification stable at 100--150 items & At 226 items, 6 of 20 models reproduce their tier in at least 90\% of half-samples \\
Unscreened invalid models produce AUROC at chance (.499--.509) & Not supported; the non-Valid models' ordinal AUROC ranges from .32 to .55 \\
Within-Valid L vs AUROC $\rho = -.80$; per-track $r$ vs AUROC $\rho = .788$ ($n = 107$) & $\rho = -.40$ ($p = .14$); per-track $\rho = .74$ ($n = 77$, five tracks) \\
L $\alpha = .921$, K $\alpha = .953$; two-factor PCA (94.6\%); L--K $r = .82$; F--Fp $r = .99$; L and F independent ($r = -.10$) & L $\alpha = .64$, K $\alpha = .88$ (17 models with complete tracks); PCA removed (exact linear dependence among the indices); L--K $r = .70$; F--Fp $r = .98$ (near-duplicates by construction); L--F $r = -.65$ \\
Valid models' item sensitivity individually significant in 13--15 of 14--17 & Under v0.3 every Valid model is significant by definition \\
Retrospective and prospective probes disagree in 12 of 20 models & Rests on the free-text track; withdrawn pending regrading \\
TRIN $\geq .95$ flags 11 models & 8 models (five keyed tracks) \\
Only R1 exceeds 0.6\% WITHDRAW+BET & Gemma 3 1B 17\%, GPT-5.4 nano 3\% of correct items \\
Synthetic always-WITHDRAW RBS $= +1.000$; eight fixed policies all correct & RBS $= 0$; replaced by 2,000-draw simulation (Table~\ref{tab:synthetic}) \\
AUROC $= .5$ is consistent with perfect inversion & Perfect inversion gives AUROC $= 0$ \\
Tier stable across L thresholds .93--.97 & Holds (no tier changes) \\
OpenAI L decreases with scale & Not monotone (.517, .694, .607) \\
Yang et al.: 1 of 11 Invalid, mean AUROC .624, $\rho(\text{L},\text{AUROC}) = +.894$; arXiv:2404.09272 & Not computed from the released data. Replaced by Section~\ref{sec:yang}; correct citation arXiv:2412.14737 \\
MMLU: battery-Invalid models become Valid; passing the binary screen implies passing the continuous one & First half holds at cut point 90. Second half is false (Gemma 3 1B, 12B, 27B) \\
R1 MMLU AUROC .768 & .806 (MMLU-500); .769 (atlas) \\
\bottomrule
\end{tabular}
\end{center}

\end{document}